\documentclass[runningheads]{llncs}

\usepackage{graphicx}
\usepackage{hyperref}
\usepackage{comment}
\usepackage{booktabs}
\usepackage{multirow}
\usepackage{url}

\begin{document}

\title{SonOpt: Sonifying Bi-objective Population-Based Optimization Algorithms
\thanks{This work was supported by the Engineering and Physical Sciences Research Council [grant number 2492452].}
}

\author{Tasos Asonitis\inst{1}\orcidID{0000-0003-2413-8752} \and Richard Allmendinger\inst{1}\orcidID{0000-0003-1236-3143} \and Matt Benatan\inst{2}\orcidID{0000-0002-8611-3596} \and Ricardo Climent\inst{1}\orcidID{0000-0002-0484-8055}}
\authorrunning{T. Asonitis et al.}

 \institute{University of Manchester, Manchester M15 6PB, United Kingdom \and Sonos Experience Limited, London SE1 3ER, United Kingdom}

\maketitle              
\begin{abstract}
We propose SonOpt, the first (open source) data sonification application for monitoring the progress of bi-objective population-based optimization algorithms during search, to facilitate algorithm understanding. SonOpt provides insights into convergence/stagnation of search, the evolution of the approximation set shape, location of recurring points in the approximation set, and population diversity. The benefits of data sonification have been shown for various non-optimization related monitoring tasks. However, very few attempts have been made in the context of optimization and their focus has been exclusively on single-objective problems. In comparison, SonOpt is designed for bi-objective optimization problems, relies on objective function values of non-dominated solutions only, and is designed with the user (listener) in mind; avoiding convolution of multiple sounds and prioritising ease of familiarizing with the system. This is achieved using two sonification paths relying on the concepts of wavetable and additive synthesis. 

This paper motivates and describes the architecture of SonOpt, and then validates SonOpt for two popular multi-objective optimization algorithms (NSGA-II and MOEA/D). Experience SonOpt yourself via \url{https://github.com/tasos-a/SonOpt-1.0}.

\keywords{sonification  \and optimization \and algorithmic behaviour \and metaheuristics \and evolutionary computation \and process monitoring \and SonOpt.}
\end{abstract}

\section{Introduction}

Sonification deals with the representation and communication of data through sound and has been an established practice for more than three decades. Blurring the lines between art and science, sonification has long been an area of interest for both musicians and scientists. The former see in it the opportunity to explore the aesthetic content of phenomena that surround us, while the latter tend to see it as a tool that can represent data efficiently and reveal hidden patterns in an intuitive way~\cite{gresham2012relationships}. Field experts tend to use sonification in tandem with visualization techniques~\cite{poguntke2008auditory}, in order to maximize the benefits of both approaches. However, sonification has proven useful as a standalone tool as well, particularly when the temporal evolution of data is an important factor of the process in question~\cite{de2007toward}.

This highlights the type of operations in which sonification really shines. For spatial representations, visualization methods generally tend to provide a concise picture. As discussed in~\cite{sawe2020using}, it is not uncommon that these methods might be proven less effective when it comes to time-series data representation. The evolution of an optimization algorithm is a process over time, and, if characterized in terms of metrics, can be thought of as a multivariate time-series. Gaining a better understanding about the behaviour of an optimization algorithm during evolution is important to study strengths and weaknesses of algorithms, and ultimately develop more efficient algorithms. While this is true for any type of algorithm, the focus of our work is on population-based multi-objective optimization algorithms (MOAs) due to the common scenario of having multiple conflicting objectives in practice~\cite{miettinen2012nonlinear}. The community has looked to theory (e.g. runtime analysis~\cite{1288055}) and visualisation-based concepts (e.g.~\cite{6777535}), including the design of visualisable test problems~\cite{9440928}, to improve algorithm understanding, but there is very little work that explores the application of sonification for this task. The goal of our work is to fill this gap, and this paper is the first attempt to sonify MOAs.

As algorithm designers and/or practitioners, we tend to evaluate the performance of optimization algorithms over the course of a few runs. As explained in ~\cite{grond2011interactive}, sonification allows us to zoom in on an individual run of the optimization algorithm, and reveal particular characteristics that can help us validate the algorithmic performance.  

Only a few approaches have been made for monitoring of optimization algorithms using sound. These will be presented briefly in the following section. Despite the existence of different methodologies, none of the currently documented implementations has focused on the evolving shape of the approximation set (objective function values of non-dominated solutions\footnote{A non-dominated solution is one which provides a suitable compromise between all objectives without degrading any of them. For a more formal definition of multi-objective concepts please refer to e.g.~\cite{miettinen2012nonlinear}.}) and the recurrence of points in the objective space. SonOpt receives as input the approximation set and maps the shape of this set to the shape of a single-cycle waveform inside a buffer. This buffer is used as a look-up table for a ramp oscillator that scans the content of the buffer repeatedly, at the rate of a user-controlled frequency. In addition, SonOpt monitors the recurrence of objective values across consecutive generations and maps the amount of recurrence of each value to the level of a different harmonic partial of a user-defined fundamental frequency. A more detailed description of the system follows on Section~\ref{SonOpt}.

The motivation behind SonOpt is to engage with the discussion on the benefits of auditory display in optimization monitoring, encourage the thinking of algorithmic behaviour in sonic terms, and offer a tool that can compliment current visualization techniques towards an intuitive multi-modal understanding of complex processes. Although not explored in this paper, an additional aspiration behind SonOpt is to explore optimization processes from an aesthetic point of view and suggest the treatment of such processes as music composition tools. The structure of this paper is divided into four sections. First we put our work in context with related literature focusing on areas such as sonification of optimization algorithms and benefits of sonification in process monitoring. Next we present the platform of SonOpt by describing the pipeline and discussing the reasons behind specific design choices. The documentation of test results is included in Section~\ref{tests}. Finally, Section~\ref{end} presents some concluding remarks and future plans.

\section{Related work}
\label{relatedwork}

The first documented approach that employed sound to provide insights in evolutionary optimization was proposed by Grond et al.~\cite{grond2011interactive} in 2011. The presented work involves the use of stacked band-pass filters to transform the sounds of keys on a computer keyboard. Grond et al. highlight the crucial role of auditory display in monitoring operations and explain how these advantages pertain to optimization processes. Their system focuses on evolutionary strategies (ES) as presented in~\cite{rechenberg1978evolutionsstrategien}. However, the developed methodology cannot be applied to optimization tasks with more than one objective.      

Tavares and Godoy~\cite{tavares2013sonification} have used sound to represent population dynamics in a Particle Swarm Optimization (PSO) algorithm~\cite{kennedy1995particle}. Although PSO has been extensively employed as a music composition tool (for example~\cite{blackwell2004self}) through the sonification of particle trajectories, Tavares and Godoy follow a different approach by sonifying specific performance metrics. Their focus is on providing information about the population behaviour --- including speed, alignment and diversity of particles within a population (or swarm) --- with particular attention to creating an aesthetically pleasing output. The sonic mappings implemented include musical notes, echo and harmonics and aim to result in evolving soundscapes. As in~\cite{grond2011interactive}, the focus is on single-objective optimization.  

Lutton et al.~\cite{lutton2015visual} apply sonification in island-based parallel setups where the optimization task has been distributed across systems to facilitate computational efficiency. Each computational node is assigned a part of a musical score, or a familiar song in audio format. The degree to which the music is reproduced faithfully --- including performance errors or glitches, depending on whether the user has opted for a musical score or audio file --- provides information about the optimization progress of individual nodes. The implemented system is targeted exclusively at parallel setups. A potential question relates to whether the correct interpretation of the introduced errors requires prior musical training, and if that could hinder the effectiveness of the designed sonification, as indicated in~\cite{neuhoff2019sonification}.  

Albeit not specifically related to optimization, a detailed account of sonification including theory, technology, applications and benefits, has been made by Hermann at al.~\cite{hermann2011sonification}. Vickers~\cite{vickers2011sonification} focuses on the benefits of sonification for peripheral process monitoring, and includes a list of systems that employed sound to monitor the running process of various programs, mainly through the triggering of sounds based on specific events. The applications mentioned in~\cite{vickers2011sonification}  do not target optimization processes. They are primarily focused on debugging run-time behaviour and most of them relate to dated programming methods. However, the variety in approaches shows that using sound to monitor algorithmic processes has been drawing attention for a long time. According to Vickers, peripheral process monitoring is when useful information about a process is being aurally displayed on the background, while a different, primary task, usually visual, is at the center of focus. SonOpt is designed precisely with this in mind. By using two separate audio streams, one that acts as a continuous sound stream and one that is triggered by events, SonOpt allows the user to focus on a different task while indirectly monitoring the progress of the optimization algorithm. As SonOpt does not aspire to replace visualization methods, but to work in tandem with them, users are encouraged to engage in multi-modal monitoring by combining visualization of various optimization aspects, with the sonification provided by SonOpt. Hildebrandt et. al~\cite{hildebrandt2016continuous} carried out an experiment in which participants performed significantly better at a peripheral process monitoring task when continuous sonification and visual cues of the process in question were used in parallel. In particular, as shown in~\cite{poguntke2008auditory}, field experts demonstrate preference towards the combination of continuous sonification and visual displays when it comes to process monitoring, instead of visual or audio only conditions. The experimental work of Axon et. al.~\cite{axon2020data} also points towards benefits of adding sonification to visual representations, specifically targeted in security related process monitoring. In~\cite{iber2021auditory} additional evidence on performance improvement in process monitoring (of physical production systems) via sonification is provided. 

The term \textit{sonification} is used throughout this text because it is a recognized and all-encompassing term that relates to auditory display. However the methodology used in SonOpt further pertains to a subbranch of sonification, referred to in the dedicated literature as \textit{audification}. Kramer~\cite{kramer2000auditory} writes that audification is ``the process of directly playing back the data samples''. As elaborated by Dombois and Eckel~\cite{dombois2011sonification}, audification can provide a successful choice of auditory display when datasets feature specific properties. The requirements include: the size of the dataset, whether the array of data samples creates a wave-like signal, the complexity of the signal, and the potential existence of subtle changes. The approximation set obtained by a MOA, which functions as the input to SonOpt, fulfills these properties. As will be evidenced in the next section, audification is applied on the first of the two sound generation paths of SonOpt.

To put this work into context with the wider literature, it is worthwhile mentioning a few more related research streams. In particular, the premises and aims in the work of Ochoa, et al.~\cite{ochoa2021search} on Search Trajectory Networks (STNs), albeit focused on the visual domain through the use of graphs, share a common ground with the motivation behind SonOpt. Both projects aim to assist with the understanding of algorithmic behaviour. The results of STNs are particularly interesting from an aesthetic point of view, showing that care has been taken on that aspect as well as improving understanding of search behaviour. 

Schuller et al.~\cite{10.1145/3462244.3479879} argue that sonification is a promising and user-friendly alternative to visualization, towards explainable Artificial Intelligence (XAI). Lyu et al.~\cite{DBLP:journals/corr/abs-2109-15193} apply an interactive virtual environment that allows real time tweaking of a neural network's hyperparameters by providing users with auditory feedback on the impact of their choices. Finally,~\cite{AliMAAJ19} highlights the shortcomings of visual representations for non-trained and visually impaired individuals, and suggest sonification as one of the prevalent alternatives.

\section{Methodology and system overview}\label{SonOpt}
SonOpt is an application developed using the visual programming environment Max/MSP~\cite{puckette1988patcher}. Max/MSP has been the choice of musicians, sonic artists and engineers, for developing audio-related programs which benefit from a certain degree of customization. We chose Max/MSP for the immediacy of design, and its popularity among sonification researchers. This section will describe the system and discuss some of the design decisions made during its development.

SonOpt can work with any population-based search algorithm. This is possible as it only needs to receive the objective function values of non-dominated solutions at each generation of the algorithm (e.g. in the case of steady state algorithms); for ease of presentation, we will use the term \textit{generation} knowing that the concepts can also be applied to non-generational (e.g. steady state) algorithms. Each generation produces a collection of objective values that are passed to SonOpt through the use of the Open Sound Control protocol (OSC)~\cite{freed1997open}. This is to ensure quick and precise communication between the algorithm and Max/MSP. The algorithms and the optimization problems presented in this paper come from pymoo~\cite{pymoo}, a Python library tailored to MOAs. Pymoo was chosen as it allows for simple swapping between optimization algorithms and problems, facilitating direct comparisons of a variety of setups. Another key motivating factor is pymoo's popularity within the optimization community: having a familiar tool at the center of the work will make it easier for optimization researchers to engage with SonOpt. Of course, users can replace pymoo with their own code, for example, of customized algorithms. 

\begin{figure}[!t]
\includegraphics[width=\textwidth]{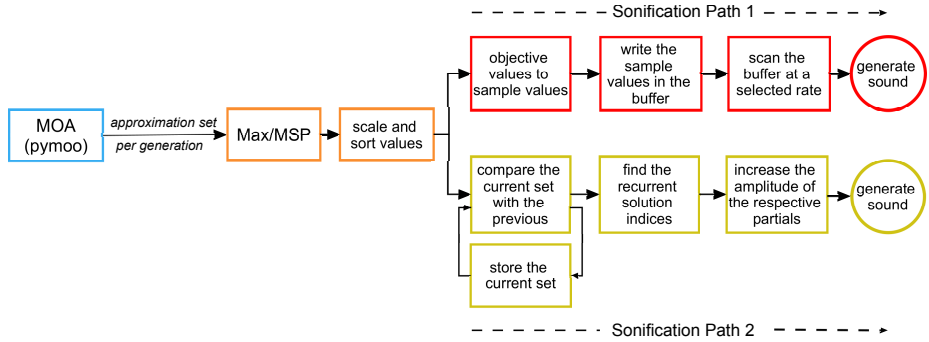}
\caption{The pipeline of SonOpt.} \label{Pipeline}
\end{figure}

At the moment, SonOpt is set up to work for bi-objective optimization problems only; WOLOG it is assumed that both objectives are to be minimized. Since SonOpt's sonification relies on the objective function values only (at least for now), a 2-dimensional matrix is expected to be passed on from pymoo to Max at each generation to represent the approximation set at that generation. Once received, Max scales the objective values to be within [0,1], and then sorts the matrix, from the minimum all the way to the maximum value of objective one --- hence from the max to the min value of objective two. This produces an ordered sequence of points comprising the 2D approximation set. Subsequently, the set is forwarded to two separate sonification paths (Fig.~\ref{Pipeline}). 

\begin{figure}[t!]
\includegraphics[width=\textwidth]{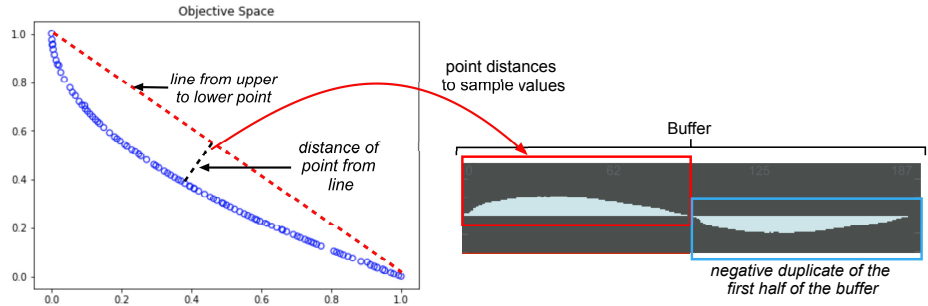}
\caption{Sonification path 1: mapping of the (scaled) approximation set shape to sample values. Algorithm: NSGA-II. Optimization task: ZDT1.} \label{Fig.2}
\end{figure}

\vspace{+1mm}
\noindent\textbf{Sonification path 1\quad}The first path is calculating a straight line from the point that minimizes objective one to the point that minimizes objective two (red dashed line in the left plot of Fig.~\ref{Fig.2}). It then calculates the distance of each point of the set from this line. The distances are then scaled and mapped to the sample values of a buffer. The buffer needs to be at least $2N$ samples big, where $N$ is the number of maximum points in the approximation set obtained per generation. The content of the buffer then looks like a horizontal display of the approximation set (right plot of Fig.~\ref{Fig.2}). In order to generate a bipolar audio signal, the incoming values are converted to negatives and are added to the buffer content. This buffer is used as a look-up table of a wavetable oscillator~\cite{bristow-johnson1996wavetable} that scans the buffer content at the rate of a user-selected frequency. This way the shape of the approximation set is treated like a single-cycle waveform, which gets repeated periodically. The resulting sound is a direct translation of the shape of the front, and it evolves according to the update of the shape at each generation. It is worth noting here that although the buffer size is fixed, the size of the readable buffer content is not; it depends on the number of objective values the algorithm is generating during the current generation. So even if the buffer looks like it has leftover content from previous generations where the number of values was bigger than the current one, the wavetable oscillator only reads the part of the buffer that includes the content generated during the current generation. If, for example, in generation 46 the algorithm generates 90 values, the buffer content will be 180 samples long (90 points in the set converted to sample values plus the negative duplicates of these values). Now, if in generation 47 the algorithm generates 70 values, this means that the buffer content will still be 180 samples long, however only the first 140 of them will contribute to the sonic output, since 40 samples of the buffer content were left over from the previous generation. 

This approach offers a sonic portrayal of the approximation set. Characteristics like discontinuity of the front, lack of diversity and unsuccessful convergence translate to harsher, more harmonically rich sounds, while convex, concave and generally continuous fronts result in more well-rounded, ``bassy'' tones that approach the sonic characteristics of a sine wave. The loudness of the sound depends on the curvature of the front: the curvier the front, the louder the sound and vice versa. Detailed examples follow on the next section.  

\vspace{+1mm}
\noindent\textbf{Sonification path 2\quad}
The second path examines the recurrence of objective function values across successive generations. This is done by comparing the values within the received approximation set of the current generation with those within the set of the previous generation. If any recurrence is spotted, SonOpt finds the index position of the recurrent value within the set of the current generation (as aforementioned, the operations are taking place on sorted approximation sets, see Fig.~\ref{Pipeline}). The recurrent indices are then mapped to the partials of a user-defined fundamental frequency. The max number of available partials is equal to the max number of objective function values on a single generation. Depending on the position indices of the recurrent solutions across the set, the amplitude of the corresponding partials is increased. A recurrence on the first point of the approximation set (this would translate on the point located on the upper left corner of the front) would result in a raise in the amplitude of the first partial and so on. When there is no recurrence, this path does not generate sound.

\begin{figure}[t!]
\includegraphics[width=\textwidth]{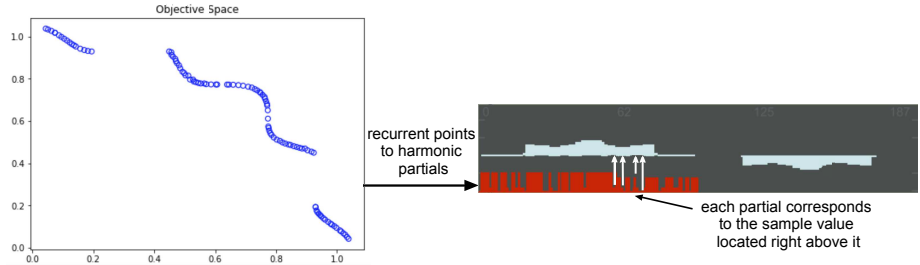}
\caption{Sonification path 2: mapping of the recurrent solutions to harmonic partials. Algorithm: NSGA-II. Optimization task: Tanaka~\cite{537993}. The amplitude of each partial, shown by the height of the corresponding red bar, shows the recurrence amount of that objective value: higher amplitude means more successive generations in which the value showed up in the approximation set.} \label{fig3}
\end{figure}

Consequently, the timbre of the resulting sound depends on the location of the recurrent values across the approximation set. When recurrence is taking place across the entire set, the outcome is rich spectrum, with many harmonic partials contributing to the sonic result, akin to the principles of additive synthesis~\cite{sasaki1980simple}. When recurrence is happening on a single value, or is concentrated on a particular area of the set, this results in more focused spectrum, involving low, mid or high range frequency components, depending on whether the recurrent area of the front is on the high left corner, middle, or low right corner respectively. The sonic outcome thus provides information not just on when recurrence is happening, but also on which part of the approximation set it is taking place. When dealing with high number of objective function values, sound can provide such information in a quicker and more intuitive way than visual representations.

The two sonification paths usually produce opposite results as the optimization progresses. Sonification path 1 tends to produce harsher sounds with quick-shifting timbre in the beginning of the run and then gradually settle on smoother tones with less harmonics, depending, of course, on the shape of the approximation set. Sonification path 2 usually starts with simple tones and tends to generate more complex sounds after sonification path 1 has reached a static timbre, i.e. the algorithm has approximated the objective space shape to some extent. This means that generally, sound intensity progressively decays in path~1, while it progressively intensifies in path 2. SonOpt was designed this way in order to prevent overlapping conveying of information that can eventually become overwhelming. Furthermore, an option is given to the users to change the volume of each path individually. SonOpt is geared towards sound synthesis, as it generates sound according to the input data. An additional reason behind this aesthetic choice is that we wanted to minimize the need for a musical background by the user. By using raw sound, one can make observations on the evolution of data intuitively, by focusing on accessible sonic qualities like loudness, harshness, low versus high frequencies etc. without needing musical training.    

\section{Experimental study}
\label{tests}

In this section we demonstrate the application of SonOpt in multi-objective optimization. A manual on how to use SonOpt and to familiarize the user with the range of sounds and their interpretation (i.e. optimization scenarios portrayed) will be released in due course. 

\subsection{Experimental setup}
\noindent \textbf{MOA settings\quad} For the presented tests, two well-known population-based MOAs were chosen: NSGA-II (Non-dominated Sorting Genetic Algorithm)~\cite{deb2002fast} and MOEA/D (Multi-Objective Evolutionary Algorithm based on Decomposition)~\cite{zhang2007moea}. The reason for selecting these two algorithms is their popularity, differences in working principles --- Pareto-dominance vs decomposition-based ---, and their availability in pymoo.\footnote{pymoo does not feature an indicator-based MOA, which we would have like to experiment with too.} As stated before, the choice of algorithm can include any population-based MOA. It is important to note that our goal here is not to discuss why one algorithm performs better than another and/or determine the best algorithm for a given problem; instead we want to demonstrate the responsiveness of SonOpt to different algorithmic behaviours and discuss insights SonOpt can provide. Consequently, we use default operators and parameter settings for our two MOAs as used in pymoo; the only parameters that need setting was the population size of NSGA-II and the number of reference vectors in MOEA/D; both parameters were set to 100 for ease of demonstration of SonOpt's capabilities. The algorithms were run on several bi-objective test problems available in pymoo. Here we show and discuss results obtained for ZDT1, ZDT4~\cite{zitzler2000comparison}, and Kursawe~\cite{kursawe1990variant} as they demonstrate SonOpt functionalities well. Importantly, video recordings of SonOpt in action can be watched at \url{https://tinyurl.com/sonopt} including also for additional problems. 

\vspace{+1mm}
\noindent \textbf{Sonification settings\quad} Table~\ref{SonOpt_Params} shows the parameters used by SonOpt for the experimental study. The setting of these parameters was determined in preliminary experimentation and driven by clarity of the sonic outputs. \textit{Sample value scaling} only affects sonification path 1 and determines how the input values are scaled in order to be converted to audible audio sample values. This greatly depends on the overall values of the approximation set. If the set includes values very close to 0, then the sample value scaling needs to be higher and vice versa. The parameter can be updated dynamically, so the user can increase or reduce the amplitude of the resulting waveform during the run, as desired. The sample value scaling has been kept steady for the examples presented in the images below. However, in the video demonstrations, the scaling value is updated when the sample values are very small, in order to make the sound characteristics of the approximation sets audible. This is happening in cases where the curvature of the set is very small. The value change has been documented. \textit{Buffer size} sets the size of the buffer in samples. For both sonification paths this needs to be at least $2N$ samples, where $N$ is the number of maximum points in the approximation set obtained per generation. \textit{Number of sliders} applies only to the second path. It refers to the number of harmonic partials that the generated sound can have. It needs to be set to, at least, $N$. \textit{Oscillator frequency} sets the frequency of the scanning oscillator in path 1 and the fundamental frequency of the harmonic oscillator in path 2. \textit{Overall amplitude} refers to the sound volume of path 1 and 2. We suggest starting with a low value and gradually increasing it, since sound bursts can be expected particularly in path 2. Finally, \textit{number of instances} controls the way Max/MSP deals with polyphony. This also needs to be set at least to $N$.

\vspace{+1mm}
\noindent \textbf{Media to gain insights into algorithmic behaviours\quad} To facilitate algorithmic understanding, SonOpt outputs sound, as an algorithm traverses through the search space (see Section~\ref{SonOpt}). We convey this to the readers via screenshots of accompanying visuals and recorded videos of SonOpt in action. It is important to mention that in the provided experiments the algorithms ran twice, once to generate the images and once to generate the video documentation, therefore variations in the optimization progress between the two might occur occasionally. Needless to say, the best way of experiencing SonOpt is by actually using it!

\begin{table}[t!]
\centering
\caption{SonOpt parameters as used in the experimental study.}
\begin{tabular}{ ccc } 
  \toprule
  SonOpt parameters & Sonification path 1 & Sonification path 2 \\ 
  \midrule
  Sample value scaling & 500 & n/a \\ 
  Buffer size & 202 & 256 \\
  Numbers of sliders & n/a & 100 \\
  Oscillator frequency & 80 Hz & 80 Hz \\
  Overall amplitude & 0.3 & 0.075   \\
  Number of instances & 100 & 100  \\
  \bottomrule
\end{tabular}
\label{SonOpt_Params}
\end{table}

\subsection{Experimental analysis}
\begin{figure}[t!]
\includegraphics[width=\textwidth]{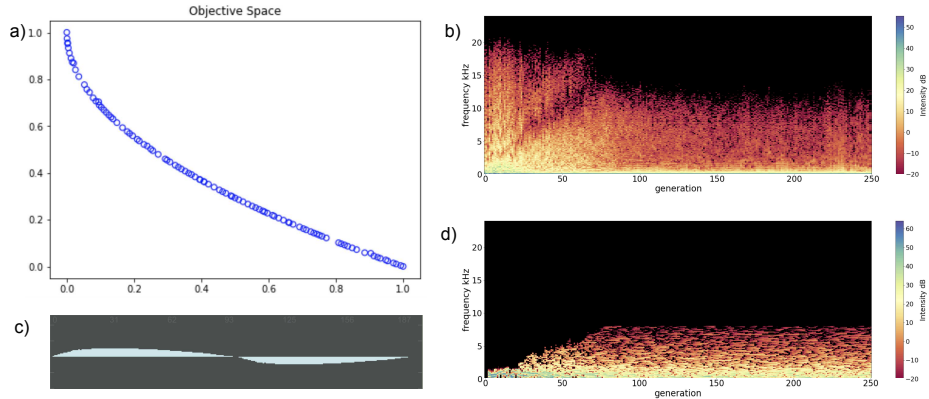}
\caption{Algorithm: NSGA-II. Optimization task: ZDT1. (a): approximation set obtained after 250 generations. (b): sonogram of sonification path 1 across 250 generations. (c): buffer contains on the 250th generation. (d): sonogram of sonification path 2 across 250 generations.} \label{NSGAIIzdt1}
\end{figure}

\begin{figure}[t!]
\includegraphics[width=\textwidth]{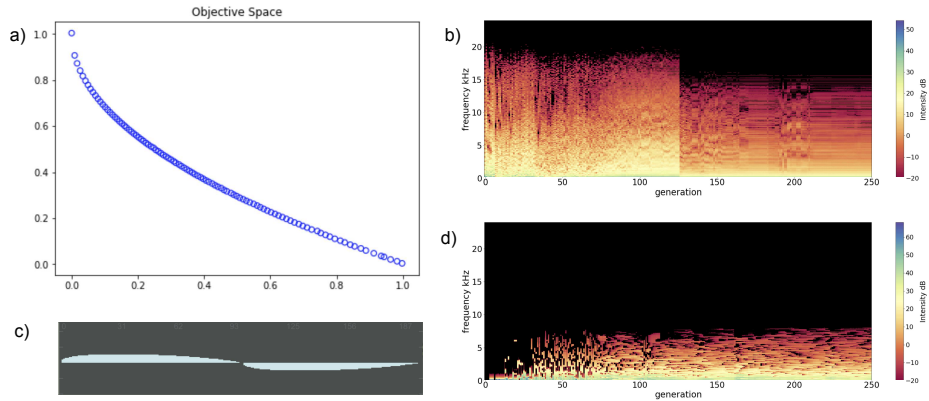}
\caption{Algorithm: MOEA/D. Optimization task: ZDT1. (a),(b),(c) and (d) as above} \label{MOEADzdt1}
\end{figure}

\noindent \textbf{SonOpt on a convex bi-objective problem\quad} Figs.~\ref{NSGAIIzdt1} and~\ref{MOEADzdt1} show the two MOAs applied to ZDT1. This is a good example of how sonification path 1 can provide insights into a run of different algorithms. In the case of MOEA/D, sonification path 1 generates a sound with noticeable harmonic content for approximately half the duration of the run. This means it did not manage to converge as quickly and effectively compared to NSGA-II, which produced a simpler, more ``punchy'' sound, approaching that of a sinetone. This is supported by looking at the obtained fronts. Both algorithms reach the convex shape, but MOEA/D presents sporadic ``bumps'' and a single solution that seems to be slightly distant from the rest of the set. This affects the resulted waveform of sonification path 1, creating a buzzier tone in comparison with NSGA-II. ZDT1 produces excessive objective value recurrence in both algorithms. Indeed, path 2 generates a waveform with gradually increasing harmonic content, as can be witnessed in Fig.~\ref{NSGAIIzdt1} (d). In the case of NSGA-II, this reflects the increasing number of non-dominated solutions obtained from generation 1 through, approximately, generation 75.

\vspace{3mm}

\begin{figure}[t!]
\includegraphics[width=\textwidth]{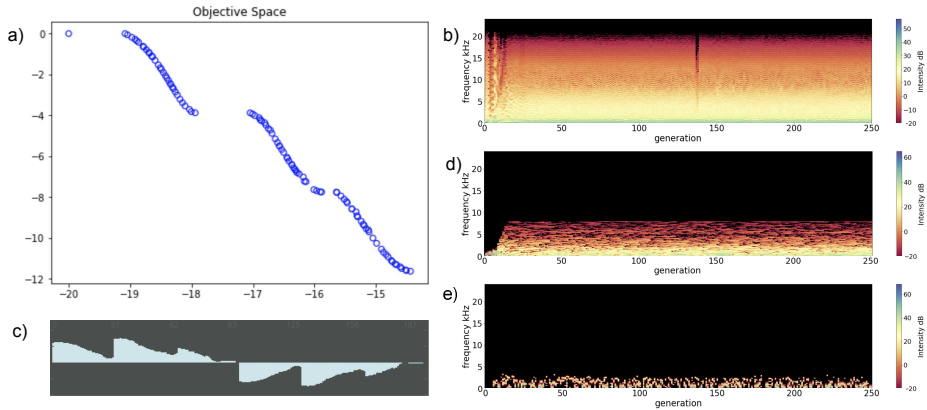}
\caption{Algorithm: NSGA-II. Optimization task: Kursawe. (a), (b), (c), (d) as above. (e): sonogram of sonification path 2, with the objective value recurrence filtered to $10\%$. } \label{NSGAIIkursawe}
\end{figure}

\begin{figure}[t!]
\includegraphics[width=\textwidth]{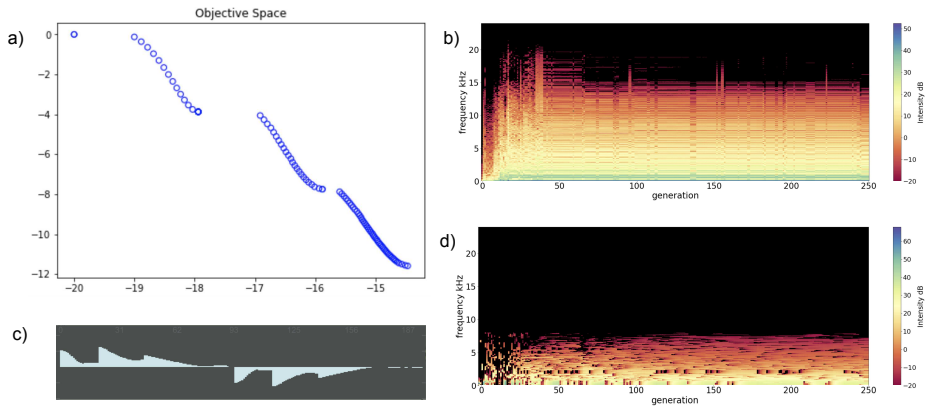}
\caption{Algorithm: MOEA/D. Optimization task: Kursawe.}  \label{MOEADkursawe}
\end{figure}

\vspace{+1mm}
\textbf{SonOpt on a discontinuous bi-objective problem\quad}In Figs.~\ref{NSGAIIkursawe} and~\ref{MOEADkursawe} we can see how SonOpt responds to NSGA-II and MOEA/D when applied to Kursawe. In this case, sonification path 1 not only provides information regarding the optimization progress, but also picks up the different behaviour of the two algorithms. For NSGA-II, path 1 generates a ``buzzy'' waveform with a quickly reduced harmonic content during the first 15 to 20 generations. For MOEA/D, however, path 1 produces a more ``jagged'' sound, with discrete, abrupt changes in the harmonic content. This reveals that the approximation set evolved in a similar, almost step-like manner until it reached the Pareto front shape. This type of behaviour is expected since NSGA-II is based on non-dominated value sorting while MOEA/D is based on decomposition. Path 2 provides additional insight. For both algorithms, path 2 produces a granular sound during the first 20 to 30 generations. This sound gets progressively substituted by a more continuous waveform, where individual tones are less discernible (see, for example, Fig.~\ref{MOEADkursawe}~(d) where the fractured harmonics in the beginning turn gradually into longer streams). This is attributed to the optimization progress. As the approximation set improves, the objective values repeat successively for more generations, a possible sign that the algorithm has converged, or got stuck at a local optimum. In Fig.~\ref{NSGAIIkursawe}~(e) we added a sonogram of path 2 after applying a filter to the recurrent values of each generation that leaves only a $10\%$ of these values come through to path 2. This is provided as a demonstration of how path 2 sounds like when the objective value recurrence is less than in the problems/algorithms we examine here. For example, this could apply to algorithms that use customized diversity-maintenance mechanisms.

\begin{figure}[t!]
\includegraphics[width=\textwidth]{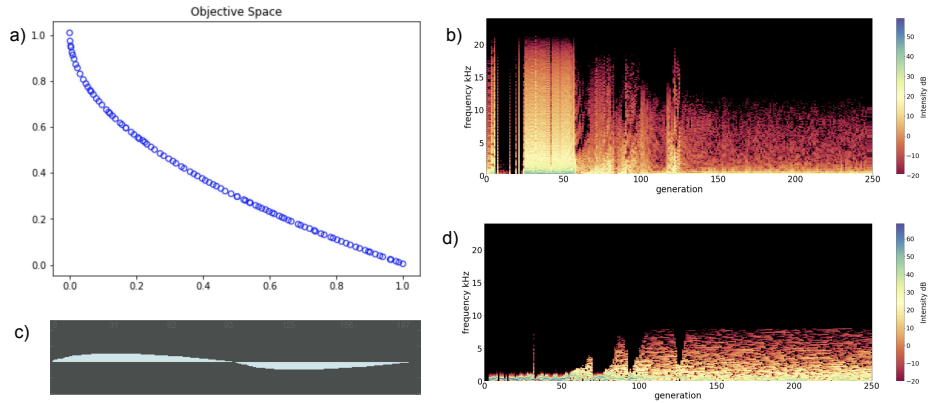}
\caption{Algorithm: NSGA-II. Optimization task: ZDT4.} \label{NSGAIIzdt4}
\end{figure}

\begin{figure}
\includegraphics[width=\textwidth]{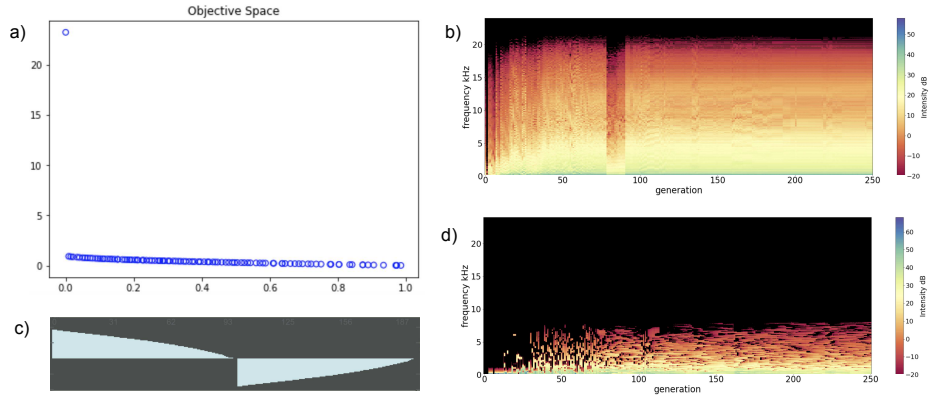}
\caption{Algorithm: MOEA/D. Optimization task: ZDT4.} \label{MOEADzdt4}
\end{figure}

\vspace{+1mm}
\noindent \textbf{SonOpt on a convex multimodal bi-objective problem\quad}
An effective example of paths 1 and 2 can be observed when applying NSGA-II to ZDT4 (Fig.~\ref{NSGAIIzdt4}). The algorithm manages to approach the Pareto front shape halfway through the optimization. Until that point the shape of the approximation set is coming together and falling apart, as can be witnessed in the video. The sound of path 1 transitions through waveforms with strong harmonic content and high amplitude (sound ``bursts''), since the produced curvatures are steep. As NSGA-II manages to approximate the front, the harmonics and overall loudness are reduced, and the simpler sonic character of a sinetone emerges. Path 2 transitions from mid-range tones to tones with broader harmonic content, as can also be witnessed from the slopes in Fig.~\ref{NSGAIIzdt4}~(d). This is due to ZDT4 presenting many local optima, posing difficulties during the optimization. SonOpt is able to pick up this information and convey it through sound. The displayed run of MOEA/D on ZDT4 produced an outlier on the upper part of the front (Fig.~\ref{MOEADzdt4}~(a)). SonOpt is very sensitive to this type of behaviour. This can be noticed by comparing the obtained sound of MOEA/D and NSGA-II after 250 generations. The outlier in MOEA/D affects significantly the shape of the waveform, producing a sound close to a ``sawtooth'' wave.  

\vspace{+1mm}
\textbf{Final remarks\quad}The presented examples confirm that SonOpt can provide information on the shape of the approximation set even when visualisation is not available. This property is of use in practice, for example, for SonOpt users with impaired sight or when auditioning the optimization progress while working on a different task (multi-tasking). A quick-shifting timbre in sonification path 1 implies that the algorithm mostly performs exploration, while a more static timbre means the algorithm performs exploitation. Once the timbre has stabilized, the sonic qualities of path 1 can provide information on the shape of the front: harsh, buzzy sounds suggest a discontinuous approximation set (and thus potentially Pareto front), and softer, bassy tones allude to a convex or concave Pareto front, with loudness encapsulating the level of curvature. The characteristics of path 2 inform on possible stagnation or successful convergence. It has been stated before that the two paths work complimentary to each other and more solid conclusion can be drawn when the output of both is taken into consideration.

\section{Conclusion and future work}
\label{end}

This paper has proposed SonOpt, an (open-source) application to sonify the search process of population-based optimization algorithms for bi-objective problems. SonOpt is implemented in Max/MSP and draws live information about the search process of an algorithm (encoded in pymoo in this study) to create two parallel sound streams indicating the evolving shape of the approximation set and the recurrence of points in the objective space. Using two popular multi-objective algorithms (NSGA-II and MOEA/D) and a range of bi-objective test problems, we demonstrated that SonOpt can pick up several aspects in terms of search behaviour including convergence/stagnation, shape of the approximation set, location of recurring points in the approximation set, and population diversity. Hence, SonOpt can facilitate algorithm understanding. 

It is important to note that SonOpt is a first attempt to sonify multi-objective search so there is plenty of scope to improve the system. For instance, future work will look at extending SonOpt to problems with more than two objectives. In particular, the community could do with a better understanding of algorithms when applied to many-objective optimization problems. Further avenues of future work include a Python-based version of SonOpt (as opposed to using Max/MSP) to improve usability and increase uptake; incorporation of different/additional sonification paths (e.g. hypervolume contributions of solutions); sonifying behavioural differences across multiple algorithms simultaneously; and enhancing the perceptibility of SonOpt's output to make SonOpt even more accessible. 

Although not elaborated within this paper, the creation of SonOpt was also motivated by artistic intentions. Optimization and, generally, algorithmic processes that evolve temporally, lend themselves fittingly in music compositions. They can shape sonic narratives and musical structures in various levels, and so motivate artists to examine optimization processes from an aesthetic perspective. We believe SonOpt points towards this direction. Future work therefore includes extending the sonic palette of SonOpt and assessing its creative performance through artistic engagement with the tool.

\end{document}